# Automatic detection and counting of retina cell nuclei using deep learning

S.M. Hadi Hosseini*[a], Hao Chen[b], Monica M. Jablonski[a]

[a]Department of Ophthalmology, Hamilton Eye Institute, University of Tennessee Health Science Center, 930 Madison Ave., Memphis, TN, USA 38163;
[b]Department of Pharmacology, Addiction Science and Toxicology, University of Tennessee Health Science Center, 71 South Manassas Street, Memphis, TN, USA 38163

**ABSTRACT**

The ability to automatically detect, classify, calculate the size, number, and grade of retinal cells and other biological objects is critically important in eye disease like age-related macular degeneration (AMD). In this paper, we developed an automated tool based on deep learning technique and Mask R-CNN model to analyze large datasets of transmission electron microscopy (TEM) images and quantify retinal cells with high speed and precision. We considered three categories for outer nuclear layer (ONL) cells: live, intermediate, and pyknotic. We trained the model using a dataset of 24 samples. We then optimized the hyper-parameters using another set of 6 samples. The results of this research, after applying to the test datasets, demonstrated that our method is highly accurate for automatically detecting, categorizing, and counting cell nuclei in the ONL of the retina. Performance of our model was tested using general metrics: general mean average precision (mAP) for detection; and precision, recall, F1-score, and accuracy for categorizing and counting.

## 1. INTRODUCTION

Characterization of cellular dynamics (in normal and pathologic conditions) to discover efficacious drugs to treat eye diseases like age-related macular degeneration (AMD) is crucial[1–2]. Cell nuclei detection, grading, and counting in histological images provide quantitative information for studying changes in cells and tissues. More specifically, retinal cells quantifying will be helpful to vision researchers. However, analyzing large sets of microscopic retinal images is labor- and time-intensive. Without automatic methods, biologists and researchers need to spend long periods of time to quantify (e.g., detect, grade, count, etc.) objects manually. Furthermore, the sheer number of images and differences in selection criteria can lead to high inter- and intra-observer variability[3]. Several applications of artificial intelligence have been used in biomedical science to overcome these limitations, like automatic cancerous cell detection[3–5]. Automatic analysis of different cell distributions is a sound solution to generate accurate and robust outputs.

In recent years, the evolution of computing capabilities (concerning software and hardware) enabled the vision community to rapidly improve object detection algorithms. Most of the old object detection models are based on intelligent machine learning algorithms. However, the generation of ImageNet, an easily accessible image database, inspired many researchers in the field of image processing to build on the success of previous machine learning models[6–9]. Today, deep learning methods are used in multiple big data areas, including medical imaging[2–3, 10–19]. Concerning the object detection problem, many deep learning-based methods use convolutional neural networks (CNN) because they are useful for image data. Each image pixel is considered one feature in these models. Therefore, when using high-resolution images, there is a large feature vector on the input side. For training this model, we need a high-performance computer with a large cache-memory and a fast process unit. Recently, to solve the image processing problems, hardware components have progressed well, and high-speed and strong graphics process units (GPUs), which can process a large image dataset easily with parallel processing, are developed.

*hosseini@uthsc.edu; phone 1 901 448-1101; fax 1 901 448-5028; uthsc.edu

The **main contribution** of this work is the development of an automatic cell detection model based on deep learning techniques and Mask R-CNN object detection model[20] to analyze transmission electron microscopy (TEM) images and quantifying the cell nuclei in the outer nuclear layer (ONL) of the retina with high speed and precision. The Mask R-CNN is an extended version of the Faster R-CNN model[21] for masking the detected objects based on a region-based convolutional neural network. We used TEM images of the ONL layer of rat retina. Our model produces more features and information automatically and accurately. The Mask R-CNN is an extended version of the Faster R-CNN model[21] for masking the detected objects based on a region-based convolutional neural network. We used TEM images of the ONL layer of rat retina. Our model produces more features and information automatically and accurately. The **second contribution** of this work is the automatic categorizing of detected cell nuclei. We categorized nuclei into three classes: live, intermediate, and pyknotic. The deep learning method we used detects cell nuclei rapidly and classifies them accurately. The similarity of cell nuclei features in TEM images are between live and intermediate nuclei, or intermediate and pyknotic nuclei are difficult to distinguish manually but was detected accurately by our network. Finally, our **third contribution** is that the number of cells in each category are automatically counted by our algorithm, which is a key outcome for biological studies. Having information about the number of each category in the ONL will help our future research, such as large-scale phenotyping of retinal ganglion cell death in genetically diverse rodent populations.

## 2. DEEP LEARNING MODELS

One helpful method to overcome the limitations of retinal cell detection is using artificial intelligence (AI) and making an intelligent model to grab all retinal information automatically. In recent years, AI has progressed rapidly. One big branch of AI is machine learning (ML), which uses statistical and heuristic methods to enable the machine to mimic human behaviors and improve with experience[22]. It is divided into two major groups: Supervised learning (like regression and classification) and Unsupervised learning (like clustering). The supervised learning methods require labeled data. In ML models, sometimes computational units referred to as artificial neurons. Neurons take several signals from input data or other neurons, combine them linearly using weight parameters, and then pass the combined signals through nonlinear operations to other neurons or output layers (as a nonlinear regression model). Deep learning models are a sub-branch of ML models, which are so powerful in big data analyzation. There are more applications of deep learning in the medical field such as computer-aided diagnosis, drug discovery, object detection, medical image processing, image interpretation, image fusion, image registration, image segmentation, image-guided therapy, image retrieval, and analysis. Now deep learning has generated great interest in medical image analysis, and it is expected that it will hold a $300 million medical imaging market by 2021[5]. Deep learning models are generally divided into three main groups: deep neural network (DNN), convolutional neural network (CNN), and recurrent neural network (RNN).

### 2.1 CNN Models

Generally, CNN models are considered more specifically for image data and have very good performance in their application. However, they need a lot of labeled data for classification, segmentation, and object detection. The input of a CNN model is a stack of images with a grid-like structure, and their output is a group of feature maps, like class labels (Figure 1). The CNN models consist of three consecutive layers: convolutional layers, pooling layers, and fully connected layers. Convolutional layers contain in their terminal section nonlinear activation functions (such as ReLU, Sigmoid, Tanh, SoftMax) that detect the nonlinearity within data. Subsequently, pooling layers are used to (1) decrease the feature map size, and (2) pool the important features from big feature maps at the convolutional layer. Finally, at the end of the structure, fully connected layers are used to generate the desired outputs (e.g., class labels)[10].

The CNN models, using a mini-batch gradient descent method, can learn from the training data and find implicit relationships within the data (images). In CNN models, low-level features such as edges, lines, and dots are detected in the first (shallow) layers of the model, and high-level features such as object class labels are represented at the last (deep) layers of the model. Therefore, large models (with more layers) can produce more accurate results if the input dataset has many features. For example, one colored-image with a resolution of 4 Megapixel (2048×2048) has more than 12 million (2048×2048×3) features and requires a large CNN model, and high-performance computer to analyze it.

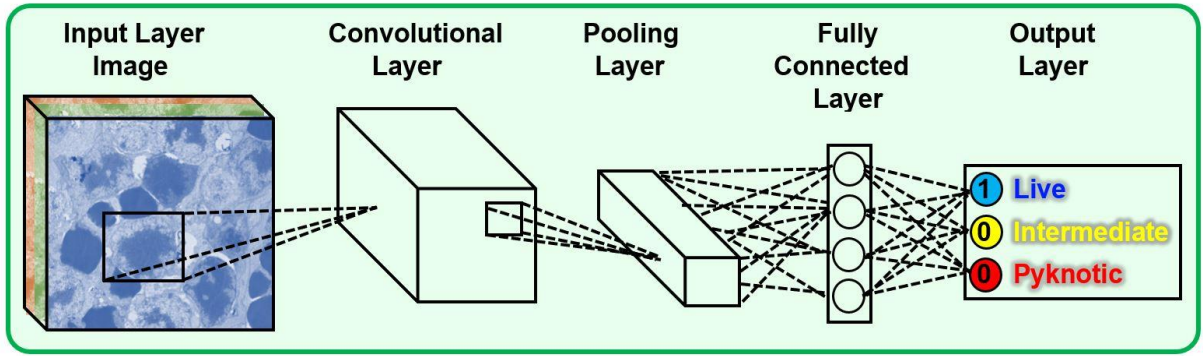

Figure 1. The structure of the CNN model.

## 3. OBJECT DETECTION MODELS

Unlike image classification problems, object detection is a multi-task learning problem. It requires the localization of the objects inside the image and then the classification of each object. In localization, the goal is to predict a bounding box/mask for each object. Then these boxes/masks are neighboring to the real bounding boxes/masks of each object in the image (ground truth) based on the confidence level of intersection over union (IoU) measurement[21]. In Figure 2, two different localization metrics based on the bounding box and mask are shown.

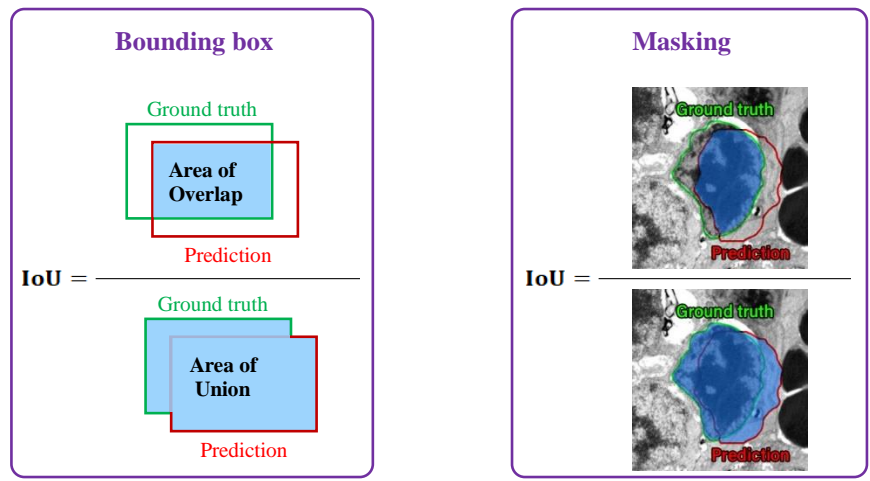

Figure 2. Two different methods for calculating IoU for localization. Localization based on the bounding box (Left). Localization based on masking (Right).

Based on different confidence levels (IoU), we can consider whether the prediction model is correct or not. For example, in Figure 3, it is shown that the predicted cell, with a threshold of 0.95 (for IoU confidence), is so fitted to the ground truth compare to other predictions. For localization, there are three different methods: (1) localization as a regression problem[23–24]; (2) building a sliding-window detector and use CNN models to fit this window with objects (Figure 4)[25]; and (3) recognition-based on developed regions[6].

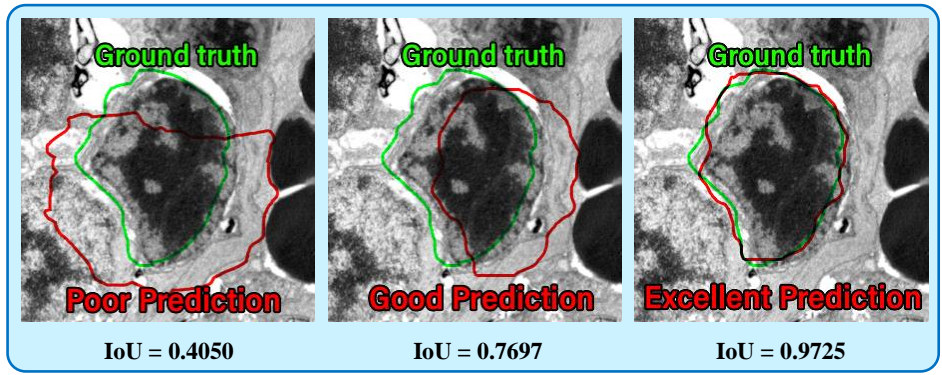

Figure 3. Three different confidential levels for localization of cell body yield three different predictions. Left: we set the confidential threshold at 0.40 and considered all predictions, even poor ones. Middle: we set the confidential threshold at 0.75 and considered the good and excellent predictions. Right: we set the confidential threshold at 0.95 and considered only the excellent predictions.

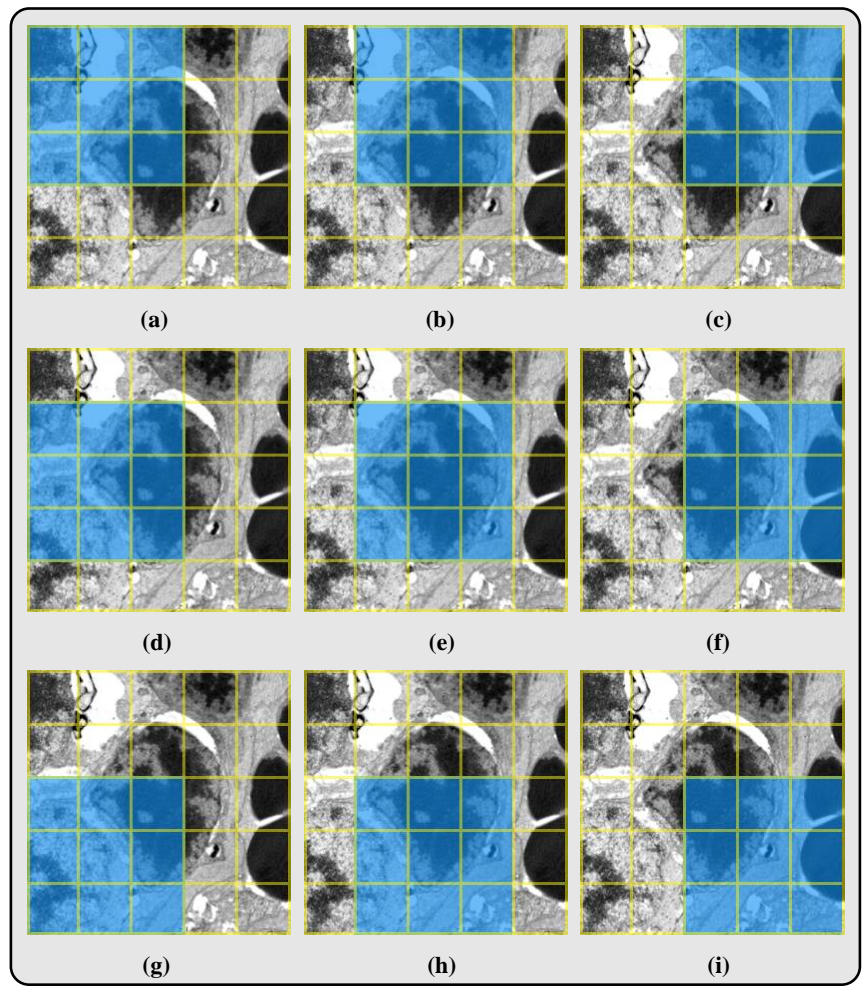

Figure 4. By sliding a window from left to right and up to down, IoU is calculated for each step. Based on the confidence threshold, the best windows are considered and merged. Thus, the predicted bounding box is generated. The best-fitted window with the middle cell is step (e).

In the third method, produce many different-size anchor boxes and fit them to the objects to find best-fitted bounding boxes for objects. So, the goal is to predict a bounding box for each object. Then these boxes are neighboring to the real bounding boxes of each object in the image (ground truth) based on the confidence level of IoU measurement (Figure 5)[23].

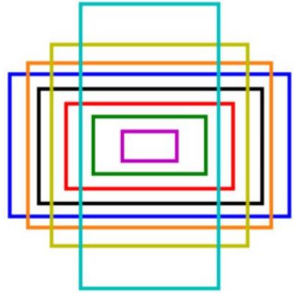

Figure 5. By sliding a window from left to right and up to down, IoU is calculated for each step. Based on the confidence threshold, the best windows are considered and merged. Thus, the predicted bounding box is generated. The best-fitted window with the middle cell is step (e).

### 3.1 R-CNN based Models

The most popular object detection methods are R-CNN, developed in 2014[6], Fast R-CNN, developed in 2015[7], Faster R-CNN, developed in 2015, YOLO, developed in 2016[25], R-FCN, developed in 2016[26], and Mask R-CNN, developed in 2017[20]. Some methods are fast like YOLO, which uses sliding windows for localization, while others have good performance in the detection of the objects, like Mask R-CNN, which uses anchor boxes for localization. In this paper, we used the Mask R-CNN model to obtain high accuracy in localizing, detecting different categories of cell nuclei, and masking them with distinct colors.

The Mask R-CNN is developed based on the Faster R-CNN model. The primary difference is that the Faster R-CNN has two outputs for each candidate object: class label and bounding box. However, in the Mask R-CNN model, a third output is added: object mask. The mask output requires the extraction of the much finer spatial layout of an object and needs pixel-to-pixel alignment between ground truth image and predicted object[20]. As shown in Figure 6, in this model there are two stages: (1) the region proposal network (RPN), which proposes a bounding box for each object, and (2) feature extraction from the considered bounding box, which explained in Fast R-CNN model[7]. In the region of interest (RoI) pooling layer, three operations are performed simultaneously: classification of the predicted object, tuning the bounding box, and binary mask prediction in pixel-level. The structure of the Mask R-CNN model capable of having backbones like well-known CNN models, such as ResNet-50, ResNet-101, and ResNet-FPN. In this study, we used ResNet-101 as a backbone for our model. Moreover, to train this model, a multi-task loss function is defined on each RoI as Equation (1).

$$L_{Total} = L_{Classification} + L_{Bounding\ Box} + L_{Masking}$$
(1)

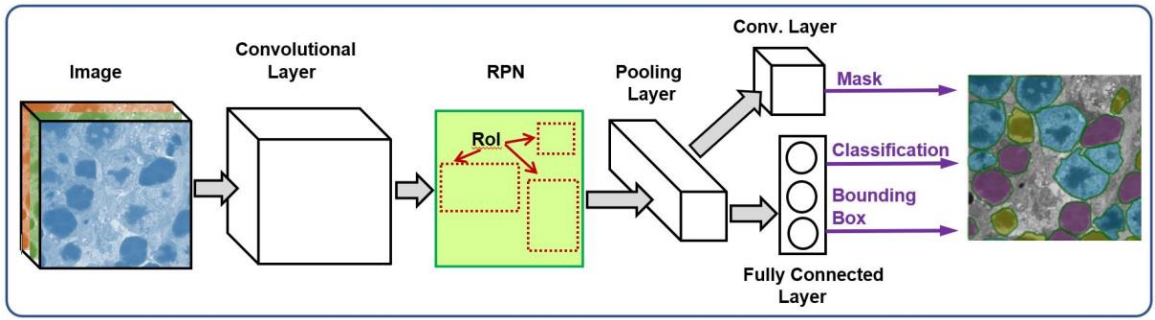

Figure 6. The structure of the Mask R-CNN model.

## 4. RETINAL ONL CELL DETECTION

The retina is the innermost, light-sensitive layer of the wall of the eye. The visual world is focused as an image on the retina, which is then translated into electrical neural impulses that are transferred to the brain via the optic nerve to create visual perception. In general, retina is divided into three important cell layers: (1) ganglion cell layer (GCL), which contains ganglion cells and displaced amacrine and bipolar cells; (2) inner nuclear layer (INL), which contains intermediate neurons and Muller cells; and (3) outer nuclear layer (ONL), which contains the cell bodies of the rod and cone photoreceptors (Figure 7). Photoreceptors contain photosensitive outer segments that are responsible for detecting light signals. These cells are, therefore, critical for preventing several types of blindness. Eye diseases, such as age-related macular degeneration (AMD), retinitis pigmentosa, and diabetic maculopathy, are the main causes of blindness. Early detection and treatment can prevent disease progression and protect vision. The information about the number of cells (in different categories) inside the ONL helps to measure the growth of blindness. Therefore, extracting important information from retinal cells leads to treating eye diseases like AMD. Cell-based information like detection and localization of cells inside each layer, classification of them into different types of cells, and counting of them in a separate category are critically important in retinal research. To detection of retinal cells, microscopic images of the retina are required. To quantifying cells inside each layer, manually counting and classifying is used by researchers. All these processes are time- and labor-intensive. Therefore, retinal researchers focused on some limited points of only the middle cross-section of the retina. For this reason, most of the whole retina is not considered, and they lost some other important information.

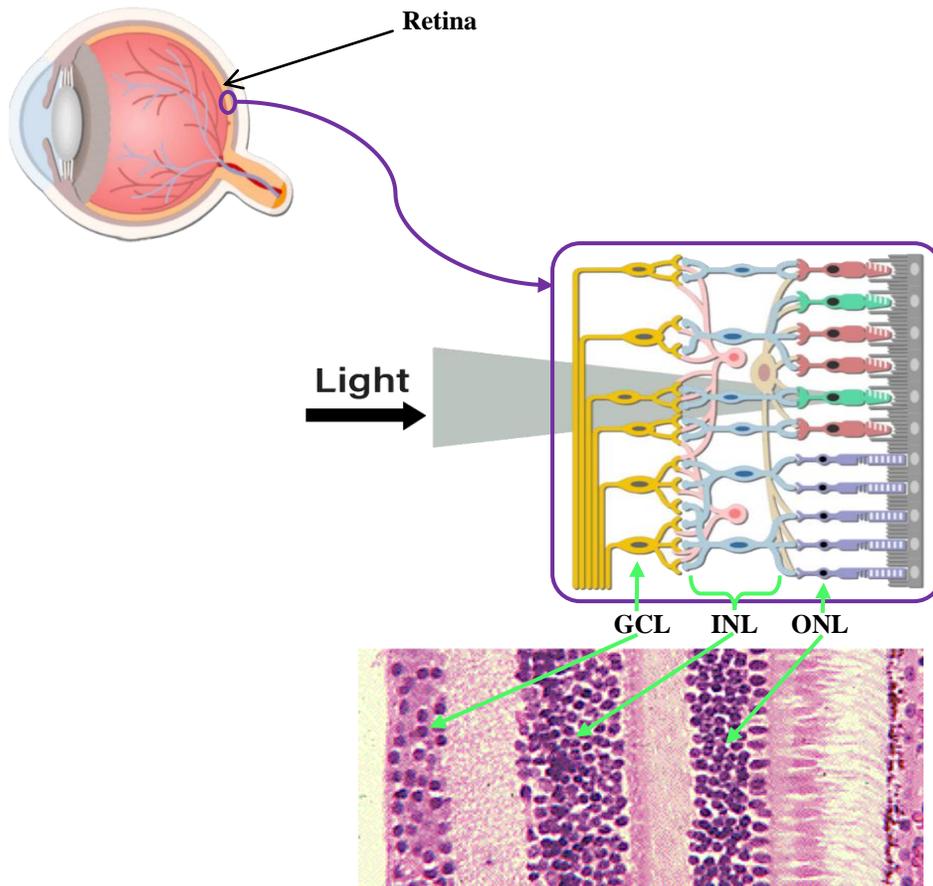

Figure 7. Eye, retina, and its cell layers: ganglion cell layer (GCL), inner nuclear layer (INL), and outer nuclear layer (ONL)[27–28].

As the first step in developing an automated method to classify cell types, we are interested in ONL cells because of the many medical conditions these cells are associated with. The goal of this study was to automatic detection of all ONL cells in the TEM images. Among powerful object detection models, we used the Mask R-CNN model for our primary structure. We obtained high accuracy in localizing, detecting, and masking the retina ONL cell nuclei.

## 5. IMPLEMENTATION

### 5.1 Dataset

In this study, we used 33 TEM images of the ONL from Royal College of Surgeons (RCS) rat retinas obtained from three experimental conditions: postnatal day 21 (P21) naïve control rats; P21 rats injected intravitreally with asialo-, tri-antennary complex-type N-glycan (NA3), a potential therapy for retinal pigment epithelial deficiency; and P21 rats injected intravitreally with phosphate buffered saline (PBS) as a negative control[29]. The RCS rat is a model of outer retinal degeneration that is characterized by marked photoreceptor death, a hallmark of which is pyknotic changes to nuclei in the ONL[30]. Images from these experimental conditions were selected because of the varying quality of nuclei of the ONL. We have 11 images from each group. All images have a high resolution of 2048×2048 pixels. Also, these images have a difference in image magnification. Some of them were captured at 3000x magnification, while others were captured at 5000x magnification. A sample of the dataset with an annotation that represents the location of cell nuclei is shown in Figure 8. We categorized and labeled the nuclei into three categories and annotated them manually using VGG Image Annotator (VIA) tool (version 2.0.8): orange line as live cells, cyan line as intermediate cells, and green lines as pyknotic cells[31]. Then, exported the annotation file as based on the COCO standard and as a JSON file for each image. Then, using the Mask R-CNN model, we masked each cell based on their category in three different colors: cyan (live cells), yellow (intermediate cells), and magenta (pyknotic cells).

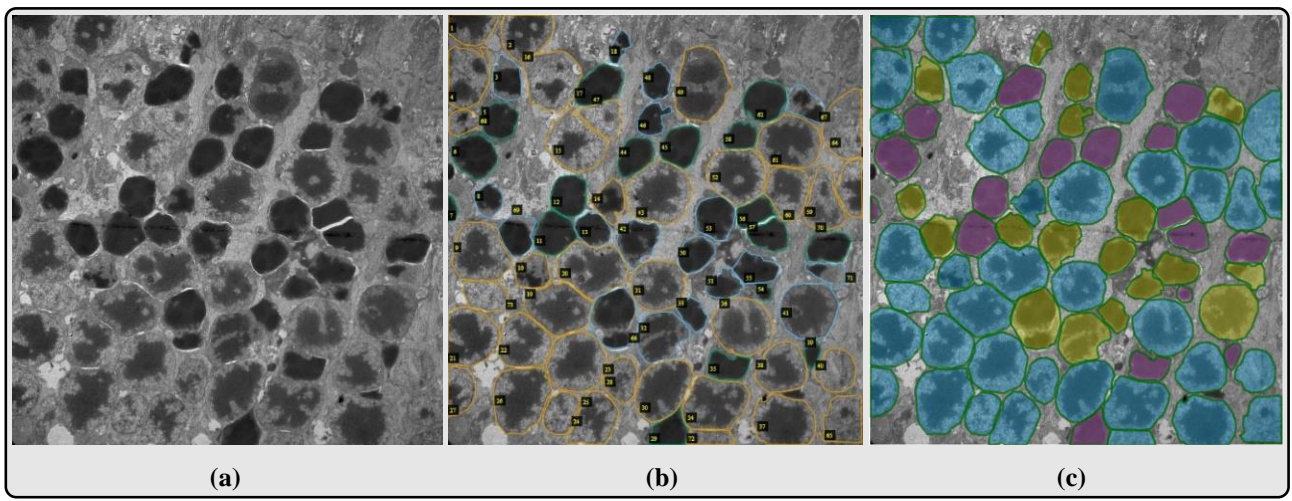

Figure 8. Transmission electron microscopy image of ONL at 5000x magnification. (a) original data, (b) image annotated and labeled by VIA tool, and (c) image masked by Mask R-CNN model in three categories: live cells (cyan), intermediate cells (yellow), and pyknotic cells (magenta).

### 5.2 Hardware

This experiment performed on a machine equipped with an Intel(R) Core(TM) i7–8700K CPU with six cores and 3.7GHz frequency, 64GB DDR2 memory, and one NVIDIA(TM) GeForce RTX 2060 GPU. The original images in our dataset have a high resolution of 2048×2048 pixels. However, our model is capable of working with lower resolution too. Then, we pre-processed the images by resizing to 512×512 pixels and adding augmentation to them. Then, we trained our model with this GPU and three images per GPU at the same time. Therefore, each training batch has three images per GPU; this is the maximum possible batch size with this GPU for our resized images, 512×512 pixels. The large mini-batch size gives high accuracy in the results, but it requires some stronger GPUs with large cache-memory. We strongly believe that our model will achieve better results if more and strong computing hardware available.

## 5.3 Training the model

We used 33 TEM images from three experimental groups (11 images per each group). Among these 33 images, we trained our model just using 24 of them (8 images per each group and in overall containing a total of 1285 cells), then validated the model and defined its structure (hyper-parameters of the model) using six images (2 images per each group with containing a total of 353 cells), and finally tested the model with remaining three images (one image per each group with containing a total of 140 cells). To compare the results of test images with the training and validation images, we also tested our model on all of 33 images.

To train the model, we considered 1200 epochs (the period of learning time that complete passes through the training and validating dataset). In each epoch, we had 100 steps for training and 100 steps for validating. In each epoch, we trained the model 100 times using the training dataset with a learning rate of 0.001 and a momentum of 0.9. Then we validated the results of our prediction 100 times using the validation dataset. Therefore, training was conducted for 120,000 iterations in each experiment. To speed up the training time, we initialized the weights of neurons of the Mask R-CNN model using the weights of the ImageNet model[32]. The implementation of the Mask R-CNN approach uses Python 3.7 language and Keras library (vs. 2.2.4)[33] with TensorFlow backend (vs. 1.13.1)[34]. Also, we used some helpful libraries in our models like NumPy (vs. 1.16.5)[35], scikit-learn (vs. 0.21.3)[36], and matplotlib (vs. 3.1.1)[37].

## 5.4 Measuring metrics

To show the performance of the Mask R-CNN model in the detection of ONL nuclei, the predicted object was compared with the ground truth image of the corresponding object. In object detection problems, there are different objects within different categories. Therefore, the localization of the classified objects is important. Subsequently, both classification and localization of the model need to be evaluated. The mean average precision (*mAP*) metric is the most common metric used in object detection problems. Therefore, to measure the performance of detection (of bounding box and mask), we used the *mAP* metric (at an IoU of 0.7).

$$mAP = mean(AP) = mean \int_0^1 P(r)dr \qquad (2)$$

where, *P* and *r* are the precision and recall of bounding box or mask detection for each object, respectively.

In general definition, the *AP* is the area under the precision-recall curve for each object. Therefore, the *mAP* refers to the mean of *AP* of all objects in each TEM image. As precision and recall are always between 0 and 1, therefore, *mAP* falls within 0 and 1 also, and the best *mAP* is 1 (highest accuracy in detection of all objects precisely). After detecting and categorizing ONL cells, the second step was accurately predicting their counts. The goal of this study was automatically predicting how many ONL cells were in each category (live, intermediate, and pyknotic) of detected ONL cells in the TEM images. To measure the performance of counting of cells in each category, four new metrics are calculated: precision, recall, F1-score (*FS*), and accuracy (*ACC*). All of these metrics are useful for classification problems and shown in Equations (3–6).

The **precision** defined as the ratio of correctly categorized cells to the total categorized cells in each category. Precision reaches its best value at 1 and worst at 0.

$$P_i = \frac{TP_i}{(TP_i + FP_i)} \quad ; \quad i = 1,2,3$$
$$P = mean(P_i) \qquad (3)$$

where, *TP* represents the correct identification of the cells' pixels, and *FP* represents the identification of the non-cells' pixels. They are classified as true positives and false positives. Also, *i* represents the number of categories.

The **recall** or sensitivity is the fraction of the relevant categorized cells that are successfully detected and defined as the ratio of correctly categorized cells to the total number of cells in each category. Recall reaches its best value at 1 and worst at 0.

$$R_i = \frac{TP_i}{(TP_i + FN_i)} \quad ; \quad i = 1,2,3$$
$$R = mean(R_i) \tag{4}$$

where, *FN* represents the wrong identification of the cells' pixels as non-cells' pixels and is classified as a false negative.

The **F1-score** with considering both precision and recall gives a chance to calculate a score for our counting and defined as the harmonic mean of precision and recall. F1-score reaches its best value at 1 (perfect precision and recall) and worst at 0.

$$FS_i = \frac{2.P_i.R_i}{(P_i + R_i)} \quad ; \quad i = 1,2,3$$
$$FS = mean(FS_i) \tag{5}$$

In numerical analysis, **accuracy** is the nearness of a calculation to the true value and defined as the ratio of the sum of correctly categorized cells and non-categorized cells to the sum of the total number of cells in each category. Like F1-score, the best value for accuracy is 1, and the worst is 0.

$$ACC_i = \frac{(TP_i + TN_i)}{(TP_i + TN_i + FP_i + FN_i)} \quad ; \quad i = 1,2,3$$
$$ACC = mean(ACC_i) \tag{6}$$

where, *TN* represents the correct identification of the non-cells' pixels and is classified as true negative.

## 6. DISCUSSION AND RESULTS

Our data demonstrate that Mask R-CNN yields results comparable in quality to manual annotation. We applied the Mask R-CNN model to the ONL nuclei for detection and categorization. For detection assessment, *mAP* metrics were used, which were calculated based on loss of bounding box, classification, and masking. The results of training the model based on loss function are shown in Figure 9. The goal of the model is decreasing the loss of prediction for bounding box, classification, and masking in the training process. Also, we considered the total loss of prediction, which is the summation of all three losses. As shown in Figure 9, the model trained well regarding all three predictions: bounding box, classification, and masking. After epoch number 600, the training loss had no significant decrease. However, in the validation dataset, after epoch number 200, the masking loss function shows an increase in the masking prediction error. It suggests that, for a new dataset, the model can detect objects very well (because bounding box error is not increased), but the correct masking of new objects needs much information.

To analyze the data, we used *mAP* metric and precision-recall curve, which are shown in Figure 10 and Table 1. According to the results of Table 1, the Mask R-CNN could predict and mask new objects with an average precision of greater than 60%.

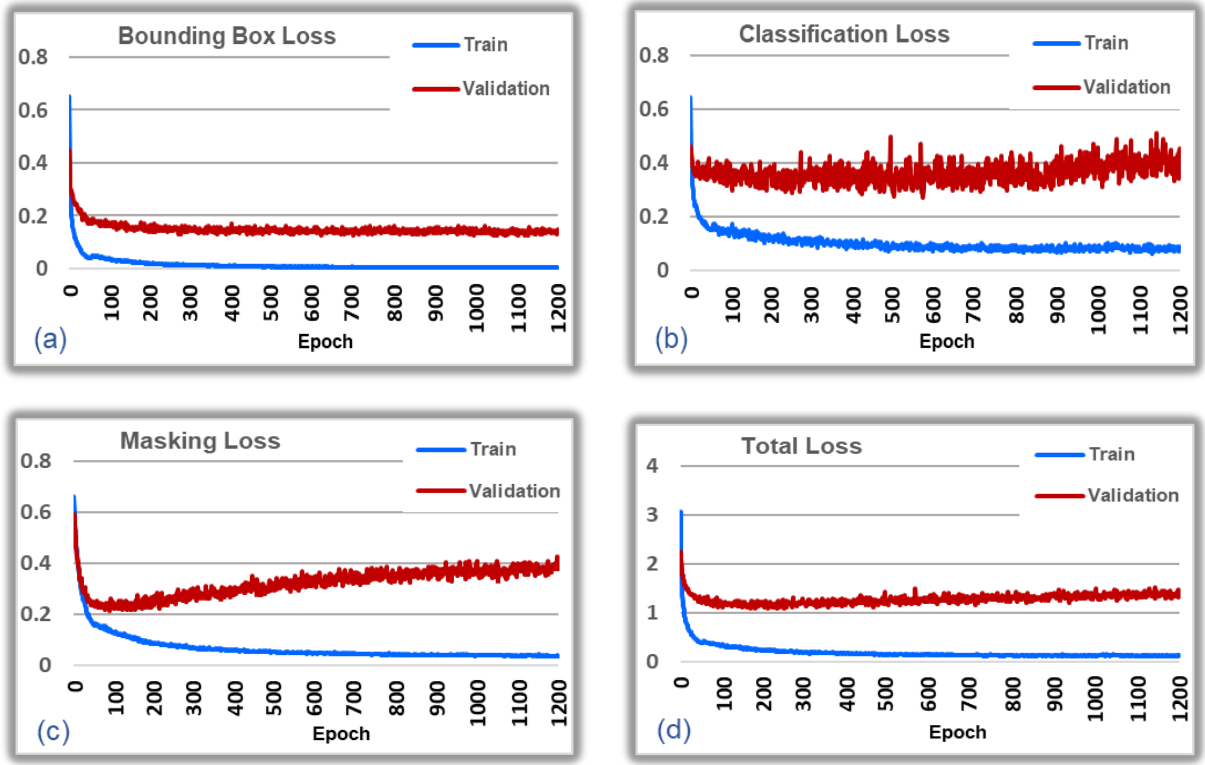

Figure 9. (**a**) Bounding box, (**b**) classification, (**c**) masking, and (**d**) total loss functions.

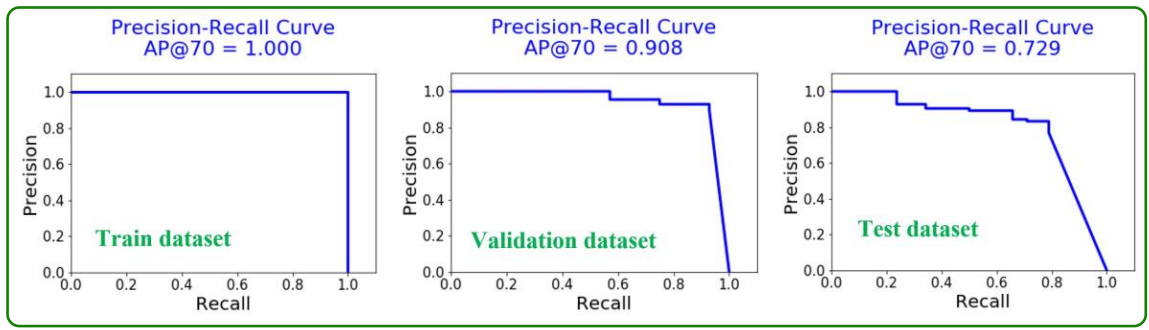

Figure 10. The precision-recall curve for one sample of each training, validation, and test datasets.

Table 1. The minimum, maximum, and average *mAP* for all datasets.

| DATASET | mAP | | |
|---|---|---|---|
| | **Minimum** | **Maximum** | **Average** |
| All 24 train data | 1.000 | 1.000 | 1.000 |
| All 6 validation data | 0.565 | 0.908 | 0.785 |
| First test data (Control) | - | - | 0.606 |
| Second test data (NA3) | - | - | 0.847 |
| Third test data (PBS) | - | - | 0.729 |
| All 3 test data | 0.606 | 0.847 | 0.727 |

At the second step, to analyze the detected objects and count them, we used precision, recall, F1-score, and accuracy, which was calculated based on the confusion matrix. All numerical analysis is shown in Figure 11 and Table 2. According to data presented in Table 2, our automatic cell counting model had an accuracy of more than 80% in cell detection and categorization. Without this automatic model, manually analyzing are time- and labor-intensive[29].

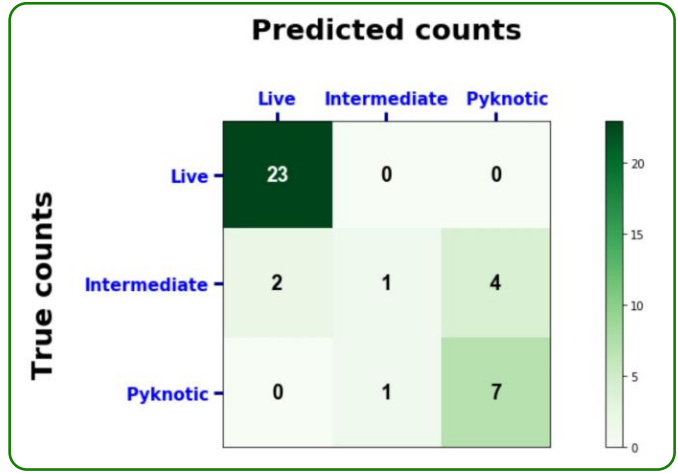

Figure 11. The confusion matrix for one of the test samples.

Table 2. The Precision, Recall, F1-score, and Accuracy for Cell Counting.

| DATASET | Counting metrics | | | |
|---|---|---|---|---|
| | Precision | Recall | F1-score | Accuracy |
| All 24 training data | 1.000 | 1.000 | 1.000 | 1.000 |
| All 6 validation data | 0.924 | 0.904 | 0.906 | 0.904 |
| First test data (Control) | 0.775 | 0.746 | 0.732 | 0.746 |
| Second test data (NA3) | 0.825 | 0.852 | 0.838 | 0.852 |
| Third test data (PBS) | 0.782 | 0.815 | 0.776 | 0.815 |
| All 3 test data | 0.794 | 0.804 | 0.782 | 0.804 |

## 7. CONCLUSION

In this study, we focused on an automatic method on detection, categorization, and counting of the cells inside microscopy images. This method is based on a deep learning algorithm and is helpful for biomedical research. In this paper, we categorized cell nuclei within TEM images of ONL of retinas from three distinct experimental conditions and counted them with an accuracy greater than 80%. Because of the advantages of the CNN deep learning models, specifically in image processing, we considered a region-based algorithm and used the Mask R-CNN model. A comparison of the results of our automatic cell detection and counting method with the results of manual annotation demonstrates that the performance of our method is high. We believe that our method can be further improved with additional training data.

## ACKNOWLEDGMENT

We acknowledge Samuel Fowler, XiangDi Wang, and Raven Simpson for manual image annotations and labeling the ONL cells. This research was supported by NIH grant number EY021200.